# Parallel Exploration via Negatively Correlated Search

**PENG YANG,** *IEEE Member*, **QI YANG, KE TANG,** *IEEE Senior Member*, **and XIN YAO,** *IEEE Fellow*

Guangdong Provincial Key Laboratory of Brain-inspired Intelligent Computation,

Department of Computer Science and Engineering,

Southern University of Science and Technology, Shenzhen 518055, China.

## Abstract

Effective exploration is a key to successful search. The recently proposed Negatively Correlated Search (NCS) tries to achieve this by parallel exploration, where a set of search processes are driven to be negatively correlated so that different promising areas of the search space can be visited simultaneously. Various applications have verified the advantages of such novel search behaviors. Nevertheless, the mathematical understandings are still lacking as the previous NCS was mostly devised by intuition. In this paper, a more principled NCS is presented, explaining that the parallel exploration is equivalent to the explicit maximization of both the population diversity and the population solution qualities, and can be optimally obtained by partially gradient descending both models with respect to each search process. For empirical assessments, the reinforcement learning tasks that largely demand exploration ability is considered. The new NCS is applied to the popular reinforcement learning problems, i.e., playing Atari games, to directly train a deep convolution network with 1.7 million connection weights in the environments with uncertain and delayed rewards. Empirical results show that the significant advantages of NCS over the compared state-of-the-art methods can be highly owed to the effective parallel exploration ability.

## Section I Introduction

Negatively Correlated Search (NCS) [1] is a recently proposed Evolutionary Algorithm (EA) [2] of iteratively searching for optimal solutions. Driven by that a properly diversified population can be more beneficial to search [3], NCS explicitly asks different subsets of the population to periodically share their probabilistic distributions so that they can cooperatively model and control the diversity of the whole population. As the probabilistic distribution actually determines how the new solutions will be sampled, NCS is featured in explicitly modeling the diversity of the next population at the current iteration. On this basis, NCS is capable of capturing the on-going interactions between successive iterations and effectively controlling the diversity of the next population, distinguishing itself from traditional EAs who only measure the diversity of sampled population [3].

Specifically, NCS explicitly divides the population into multiple exclusive sub-sets, i.e., sub-populations. The evolution of each sub-population is regarded as a separate search process and is conducted by a traditional EA for exploitation. Meanwhile, the search processes are coordinated to explore different search space by driving their probabilistic distributions to be negatively correlated. As a result, NCS has shown to perform a parallel exploration search behavior that multiple search processes are guided to search different promising areas of the search space simultaneously (see fig. 2 in [1] for illustration).

Although the basic idea of NCS has attracted increasing research interests [9]-[12] and has shown very promising performance in various real-world problems [4]-[8][24], the original instantiation of NCS [1] was mostly devised by intuition, lacking the mathematical explanations of why the negatively correlated search processes can lead to a parallel exploration and the guidance of how to optimally obtain the negatively correlated search processes.

In this paper, a mathematically principled NCS framework is proposed to address this issue. The new NCS explicitly regards the exploration and exploitation as two objectives of the general search procedure, and works by mathematically modeling and maximizing both a diversity model (for exploration) and a fitness model (for exploitation) of the next population. The diversity model measures the total negative correlations of the probabilistic distributions between pairwise search processes, and the fitness model describes the total expectation of the solution qualities that can be sampled under the probabilistic distributions. In other words, these two models respectively represent how different and how good the new solutions can be generated. By maximizing the diversity model, the search processes tend to be more negatively correlated as the "overlaps" among probabilistic distributions are getting smaller. By maximizing the fitness model, the expectation of solution qualities that can be sampled by the search processes is improved.

In practical, by employing the Natural Evolution Strategy [13] to evolve each search process, both the diversity model and the fitness model can be optimally maximized via partially gradient descending with respect to each search process. That is, each search process can independently maximize the negative correlation to the others and the expectation of sampling better solutions. On this basis, by gradient descending the two models at the same time, the resultant Negatively Correlated Natural Evolution Strategy (NCNES) is able to form a parallel exploration search behavior that different search processes will in parallel evolve to distinct yet promising areas of the search space.

To verify the effectiveness of NCNES, the reinforcement learning problem is considered for empirical studies, as it is widely acknowledged that the exploration ability has great impacts on the performance of a reinforcement learner [32]. Three popular Atari games [25] covering shooting and obstacles avoidance tasks are selected as the test instances. To play the Atari games, NCNES is required to directly train a deep convolution network with 1.7 million connection weights for optimizing the policy, which imposes great challenges to NCNES as the search space is both large-scale and highly multi-modal. Even worse, the environmental rewards are highly uncertain and heavily delayed, making the training further difficult without the help of traditional back-propagation. Empirical results have successfully shown that, NCNES can achieve significantly more scores than the state-of-the-art algorithms (including both EA-based and gradient-based solutions). Furthermore, due to the parallel exploration search behavior, it has shown that NCNES can facilitate the search more computationally efficiently with parallel computing resources.

The reminder of this paper is as follows. In Section II, the new mathematically principled NCS is presented in detail, and the weakness of the original NCS that was designed by intuition is also discussed. An instantiation of the new NCS framework, i.e., NCNES, is described in Section III. In Section IV, the effectiveness of NCNES is verified on three reinforcement learning problems by playing Atari. The conclusions are given in the Section V.

# Section II NCS for Parallel Exploration

NCS stems from re-thinking of "how does population facilitate the search?" Although it has been widely acknowledged that effective information sharing among population is the key to successful cooperative search, an open question remains what information to share and how [14]. By mimicking the cooperation in human, NCS asks the individuals in a population to have different search behaviors, so as to avoid repetitively searching a same region of the search space. Similar idea has also been adopted for ensemble learning [15]. Each search behavior is defined as how the offsprings will be sampled based on their parents, and usually can be represented as a probabilistic distribution. The mathematical correlation among distributions is utilized to statistically model the diversity among the population. As a result, by explicitly driving multiple probabilistic distributions to be negatively correlated, NCS suffices to control the diversity of the next population.

By implementing the above idea, it is necessary to instantiate a way for modeling the diversity and balancing it with exploitation. In the original NCS, such steps are mainly motivated by intuition, lacking mathematical explanations for in-depth analysis and shown to be sub-optimal. In this section, we first provide an integrated solution for these issues, and then discuss its merits over the original NCS.

**Section II.A The Mathematical Model of NCS**

Basically, the idea of NCS requires the population being exclusively grouped into $\lambda$ sub-populations, each of which is then evolved as a separate search process by a traditional EA, preferably those who sample solutions from an explicit probabilistic distribution [16]. To re-design NCS, let us start a thought game from what kind of probabilistic distribution can facilitate the search better by covering promising areas of the search space and generating new solutions therein.

It is usually straightforward to build a simple well-defined distribution like Gaussian distribution and Cauchy distribution [16]. Unfortunately, such distribution maybe incapable of capturing the complex problem characteristics like the multi-modality [17]. Usually, it is non-trivial to properly set up one complicated distribution. Similar to Gaussian Mixture Model [18], we can have multiple simple distributions instead of one complicated distribution. Another advantage of constructing multiple distributions is that we can explicitly sample different solutions therefrom for the purpose of finding multiple optima [19]. Then this problem can be turned into how to add new simple distributions to the first simple distribution. Clearly, the new distributions should be able to sample new solutions with high fitness values. Moreover, the new distributions should have fewer "overlaps" (correlations) with existing ones, so that they can be used to sample different regions of the solution space.

For clarity, let us construct the multi-distribution model from scratch. If we initially have one distribution $p(\boldsymbol{\theta}_1)$[1], there is no worry of "overlap". Thus it is only required to sample solutions with higher enough fitness values. Mathematically, this objective $\mathcal{J}$ (to be maximized) can be modeled as the expectation of fitness values[2] of the solutions $\boldsymbol{x}$ sampled from $p(\boldsymbol{\theta}_1)$ [13], shown as Eq.(1).

---

[1] $\boldsymbol{\theta}_i$ denotes the parameters of the distribution of the $i$-th search process. For simplicity, in this paper, we assume all the distributions are with the same type, e.g., Gaussian distribution, while the parameters of the distribution, e.g., mean and covariance, can be different.
[2] Without loss of generality, the maximization problem is taken for example in this paper.

$$\mathcal{J} = \int f(x)p(x|\boldsymbol{\theta}_1)\mathrm{d}x \qquad (1)$$

If we want to add a new distribution $p(\boldsymbol{\theta}_2)$ to $p(\boldsymbol{\theta}_1)$, it has to minimize the correlation between them, as well as maximizing the expected fitness values of both $p(\boldsymbol{\theta}_1)$ and $p(\boldsymbol{\theta}_2)$. For that purpose, the following Eq.(2) should be maximized.

$$\mathcal{J} = \int f(x)p(x|\boldsymbol{\theta}_1)\mathrm{d}x + \int f(x)p(x|\boldsymbol{\theta}_2)\mathrm{d}x + (-C(p(\boldsymbol{\theta}_1),p(\boldsymbol{\theta}_2)) - C(p(\boldsymbol{\theta}_2),p(\boldsymbol{\theta}_1))) \qquad (2)$$

where $C(p(\boldsymbol{\theta}_i),p(\boldsymbol{\theta}_j))$ means the correlation from the $i$-th distribution to the $j$-th distribution. Now suppose $\lambda$ distributions are considered, Eq.(2) can be readily extended to Eq.(3).

$$\mathcal{J} = \sum_{i=1}^{\lambda} \int f(x)p(x|\boldsymbol{\theta}_i)\mathrm{d}x + \sum_{i=1}^{\lambda}\sum_{j=1}^{\lambda}(-C(p(\boldsymbol{\theta}_i),p(\boldsymbol{\theta}_j))) \qquad (3)$$

By maximizing the first additive term, it says that all the distributions should be able to sample solutions with high fitness values. And by maximizing the second additive term, it means that all the distributions should be mutually negatively correlated, by which the overlaps among $\lambda$ distributions can be maximized. Given that the distributions reflect how new solutions are generated, the first additive term is able to give an expectation of how good the next population might be, and the second additive term is thus capable of modeling the diversity of the next population. On this basis, the diversity model $\mathcal{D}$ for all $\lambda$ distributions is defined as Eq.(4).

$$\mathcal{D} = \sum_{i=1}^{\lambda}\sum_{j=1}^{\lambda} -C(p(\boldsymbol{\theta}_i),p(\boldsymbol{\theta}_j)) = \sum_{i=1}^{\lambda} d(p(\boldsymbol{\theta}_i)) \qquad (4)$$

where $d(p(\boldsymbol{\theta}_i)) = \sum_{j=1}^{\lambda} -C(p(\boldsymbol{\theta}_i),p(\boldsymbol{\theta}_j))$ is the derived diversity component for the $i$-th search process. By further denoting the first additive term as $\mathcal{F}$ and its $i$-th component as $f(\boldsymbol{\theta}_i) = \int f(x)p(x|\boldsymbol{\theta}_i)\mathrm{d}x$, Eq.(3) can be re-written as Eq.(5) for clarity.

$$\mathcal{J} = \mathcal{F} + \mathcal{D} = \sum_{i=1}^{\lambda} f(\boldsymbol{\theta}_i) + \sum_{i=1}^{\lambda} d(p(\boldsymbol{\theta}_i)) \qquad (5)$$

Thus, the mathematical explanation of NCS can be expressed as maximizing the general objective $\mathcal{J}$, which turns into the maximization of both the diversity model $\mathcal{D}$ for exploration and the fitness model $\mathcal{F}$ for exploitation. It is highly desired that $\mathcal{J}$ can be maximized in parallel to eliminate the interdependencies among search processes and to enjoy the computational acceleration. Since the distribution of a search process is independent from each other by definition, one way to achieve the parallel maximization of $\mathcal{J}$ is to apply the partial gradient descent to $\mathcal{J}$ with respect to each $\boldsymbol{\theta}_i$. The gradient of Eq.(5) can be calculated as Eq.(6).

$$\nabla_{\boldsymbol{\theta}_i}\mathcal{J} = \nabla_{\boldsymbol{\theta}_i}\mathcal{F} + \nabla_{\boldsymbol{\theta}_i}\mathcal{D} = \nabla_{\boldsymbol{\theta}_i}f(\boldsymbol{\theta}_i) + \nabla_{\boldsymbol{\theta}_i}d(p(\boldsymbol{\theta}_i)), \qquad i = 1,\dots,\lambda \qquad (6)$$

Clearly, by applying the gradient descent to $\mathcal{J}$, both the diversity model $\mathcal{D}$ and the fitness model $\mathcal{F}$ of each search process can be independently maximized to enable NCS a parallel exploration search behavior, where each search process is highly likely to evolve to an un-visited promising area of the search space, respectively.

**Section II.B The New NCS Framework**

To implement Eq.(6), it is required to know how to calculate $\nabla_{\theta_i} f(\theta_i)$ and $\nabla_{\theta_i} d(p(\theta_i))$, and how to update $\theta_i$ based on them.

For $\nabla_{\theta_i} f(\theta_i)$, the work in [13] has derived the following formulation (Eq.(7)) that can be directly employed.

$$\nabla_{\theta_i} f(\theta_i) = \nabla_{\theta_i} \int f(x) p(x|\theta_i) dx = \mathbb{E}_{\theta_i}[f(x) \nabla_{\theta_i} \log p(x|\theta_i)]$$

$$\approx \frac{1}{\mu} \sum_{k=1}^{\mu} f(x_i^k) \nabla_{\theta_i} \log p(x_i^k|\theta_i) \quad (7)$$

where $x_i^k$ indicates the $k$-th solution in the $i$-th sub-population and $\mu$ is the number of the solutions in the $i$-th sub-population. For more details, please refer to [13].

To calculate $\nabla_{\theta_i} d(p(\theta_i))$, by Eq.(4), a correlation measurement $C(p(\theta_i), p(\theta_j))$ should be specified for the pair of $p(\theta_i)$ and $p(\theta_j)$. Following the original NCS, let the Bhattacharyya distance [20] be the negative correlation measurement, i.e., $C(p(\theta_i), p(\theta_j)) = -\log(\int \sqrt{p(x|\theta_i) p(x|\theta_j)} dx)$ for continuous distributions and $C(p(\theta_i), p(\theta_j)) = -\log(\sum_{x \in X} \sqrt{p(x|\theta_i) p(x|\theta_j)})$ for discrete distributions, respectively. Then $\nabla_{\theta_i} d(p(\theta_i))$ can be given as Eq.(8).

$$\nabla_{\theta_i} d(p(\theta_i)) = \sum_{j=1}^{\lambda} \nabla_{\theta_i} \log (\int \sqrt{p(x|\theta_i) p(x|\theta_j)} dx)$$

$$\nabla_{\theta_i} d(p(\theta_i)) = \sum_{j=1}^{\lambda} \nabla_{\theta_i} \log (\sum_{x \in X} \sqrt{p(x|\theta_i) p(x|\theta_j)}) \quad (8)$$

After obtaining $\nabla_{\theta_i} f(\theta_i)$ and $\nabla_{\theta_i} d(p(\theta_i))$, it is straightforward to obtain $\nabla_{\theta_i} \mathcal{J}$ by Eq.(6). Alternatively, a parameter $\varphi$ can be introduced to trade-off $\nabla_{\theta_i} f(\theta_i)$ and $\nabla_{\theta_i} d(\theta_i)$ for a subtle balance between exploitation and exploration, using Eq.(9).

$$\nabla_{\theta_i} \mathcal{J} = \nabla_{\theta_i} f(\theta_i) + \varphi \cdot \nabla_{\theta_i} d(p(\theta_i)) \quad (9)$$

Similar to standard gradient descent methods [21], the objective function $\mathcal{J}$ can be maximized by optimizing the distribution parameters with Eq.(10).

$$\theta_i = \theta_i + \eta \cdot \nabla_{\theta_i} \mathcal{J} \quad (10)$$

where $\eta$ is a step-size parameter for the gradient descending.

Based on the discussions above, the new NCS framework is listed in Algorithm I and described as follows. At the beginning stage, $\lambda$ probabilistic distributions are initialized to form a set of parallel search processes. For each iteration, the following steps are executed in parallel: 1) each $i$-th search process first generates $\mu$ candidate solutions according to its probabilistic distribution $p(\boldsymbol{\theta}_i)$ at step 6; 2) the fitness values of all $\mu$ newly generated solutions are evaluated with respect to the fitness function $f$ at step 7; 3) the gradient of the fitness model locally approximated by the $i$-th sub-population, i.e., $\nabla_{\boldsymbol{\theta}_i} f(\boldsymbol{\theta}_i)$, is calculated according to Eq.(7) at step 9; the gradient of the diversity model with respect to the $i$-th sub-population, i.e., $\nabla_{\boldsymbol{\theta}_i} d(p(\boldsymbol{\theta}_i))$, is calculated according to Eq.(8) at step 10; then the gradient of the general objective function, i.e., $\nabla_{\boldsymbol{\theta}_i} \mathcal{J}$, can be accumulated based on Eq.(9) at step 11; the general objective function $\mathcal{J}$ is thus maximized by using gradient descent method (see Eq.(10)), as shown in step 12. Finally, the best ever-found solution $\mathbf{x}^*$ that is iteratively recorded (see step 8) will be output as the result of NCS before its halting (see step 13).

---

**Algorithm I: The New NCS Framework**

| | |
|---|---|
| 1 | **Input:** $f$, $d$, $\lambda$, $\mu$, $\eta$, $\varphi$ |
| 2 | **Begin:** |
| 3 |     Initialize $\lambda$ search processes defined by probabilistic model $p(\boldsymbol{\theta}_i)$, $i = 1, \dots, \lambda$; |
| 4 | **While** stopping-criteria not met **do**: |
| 5 |     **For** each $i$-th search process: |
| 6 |         Generate $\mu$ solutions according to $p(\boldsymbol{\theta}_i)$; |
| 7 |         Evaluate the fitness of all $\mu$ generated solutions; |
| 8 |         Update $\mathbf{x}^*$ as the best solution ever found. |
| 9 |         Calculate the gradient of fitness as $\nabla_{\boldsymbol{\theta}_i} f(\boldsymbol{\theta}_i)$; |
| 10 |         Calculate the gradient of diversity as $\nabla_{\boldsymbol{\theta}_i} d(p(\boldsymbol{\theta}_i))$; |
| 11 |         $\nabla_{\boldsymbol{\theta}_i} \mathcal{J} \leftarrow \nabla_{\boldsymbol{\theta}_i} f(\boldsymbol{\theta}_i) + \varphi \cdot \nabla_{\boldsymbol{\theta}_i} d(p(\boldsymbol{\theta}_i))$; |
| 12 |         $\boldsymbol{\theta}_i \leftarrow \boldsymbol{\theta}_i + \eta \cdot \nabla_{\boldsymbol{\theta}_i} \mathcal{J}$; |
| 13 | **Output** $\mathbf{x}^*$, $f(\mathbf{x}^*)$. |

---

**Section II.C The Merits of the New NCS**

In the original NCS, there is no concept of both diversity model and fitness model. But if we look at the original NCS from this perspective, it can be found that the original NCS did not measure the expectation of qualities of unsampled solutions as the fitness model. Instead, to improve the solution qualities, it heuristically compared the fitness values of two sampled solutions for survival. This means that the original NCS cannot utilize the gradient descent method for maximizing the fitness model. Similarly, the diversity model was also maximized by such heuristic comparisons, leaving two technical issues for the original NCS, except for the unclear mathematical explanations.

To be specific, the original diversity model is basically a decentralized model. That is, the diversity of each search process was modeled individually and maximized separately. Comparatively, the new diversity model can be viewed as a centralized model since all correlations between pairwise search processes are counted together. The original diversity model of the $i$-th search process, denoted as $\bar{d}(p(\boldsymbol{\theta}_i))$, was defined as the minimum of the negative correlation between its distribution $p(\boldsymbol{\theta}_i)$ and the distributions of the other search processes, shown as Eq.(11),

$$\bar{d}(p(\boldsymbol{\theta}_i)) = \min_j\{-C(p(\boldsymbol{\theta}_i), p(\boldsymbol{\theta}_j))|j \neq i\}, \forall i,j = 1, \dots, \lambda \qquad (11)$$

To maximize each $\bar{d}(p(\boldsymbol{\theta}_i))$ of the $i$-th search process, the original NCS works by comparing the diversity of the current distribution, i.e., the parent distribution $p(\boldsymbol{\theta}_i)$ estimated from the parent sub-population, and the offspring distribution $p(\boldsymbol{\theta}_i')$ estimated from the offspring sub-population, and then selecting the larger one to update the distribution $p(\boldsymbol{\theta}_i)$ for the next iteration. In order to obtain good balance between exploration and exploitation, the fitness values are also considered during the maximization of diversity. Let $\mathbf{x}_i$ be the parents in the $i$-th search process, and $\mathbf{x}_i'$ be their offsprings. Then the heuristic comparison goes as Eq.(12),

$$\begin{cases} \text{discard } \mathbf{x}_i \text{ and } \boldsymbol{\theta}_i, & \text{if} \quad f(\mathbf{x}_i) + \varphi \cdot \bar{d}(p(\boldsymbol{\theta}_i)) < f(\mathbf{x}_i') + \varphi \cdot \bar{d}(p(\boldsymbol{\theta}_i')) \\ \text{discard } \mathbf{x}_i' \text{ and } \boldsymbol{\theta}_i', & \text{otherwise} \end{cases} \qquad (12)$$

where $\varphi \in (0, +\infty)$ is a trade-off parameter, and $f(\mathbf{x}_i)$ are the fitness values of $\mathbf{x}_i$. For more details of the original NCS, please refer to [1].

It can be clearly seen in Eq. (12) that the maximization of both the diversity and the fitness highly depends on the samplings of the candidate solutions (note that the distribution parameters $\boldsymbol{\theta}$ here are also directly estimated from the sampled solutions). However, existing sampling techniques in EAs are usually randomized and thus may involve significant noise, which may mislead the maximization of both the diversity and the fitness. Another issue is that, the above heuristic comparison suffers from the interdependencies among search processes. Specifically, by substituting Eq.(11) to Eq.(12), it can be seen that the heuristic comparison in the $i$-th search process explicitly requires the parent distribution $p(\boldsymbol{\theta}_j)$ from all the other $j$-th search processes to decide its own parent sub-population and parent distribution at the next iteration, while the heuristic comparisons in other sub-populations also require doing so. Consequently, the heuristic comparison in one search process will be interdependent from that in the others, since the parent distributions of different search processes have to be decided in sequential. Due to the above-mentioned two issues, the diversity and the fitness of each sub-population may not be maximized in parallel, possibly making the parallel exploration of NCS less effective.

Comparatively, in the new NCS, it is no longer needed to compete the exact values of the fitness and diversity pairwise between the parent and offspring sub-populations for survival, as the gradient descent mathematically provides the optimal direction for maximizing both the fitness models and diversity models. On this basis, the random noise of samplings and the interdependencies among sub-populations introduced by the original heuristic comparisons are avoided. As a result, the proposed new NCS framework has successfully addressed the two technical issues of the original NCS, and brings a much clearer explanation to the idea of NCS.

## Section III Negatively Correlated Natural Evolution Strategies

To instantiate the new NCS framework, the type of probabilistic distribution $p(\boldsymbol{\theta}_i)$ should be specified. In this paper, the Gaussian distribution is employed, i.e., $p(\boldsymbol{\theta}_i) = \mathcal{N}(\boldsymbol{m}_i, \boldsymbol{\Sigma}_i)$. The underlying reason is three-folds: 1) the Gaussian distribution is the most commonly used distribution in search [16]; 2) by using the Gaussian distribution, $\nabla_{\boldsymbol{\theta}_i} f(\boldsymbol{\theta}_i)$ has an analytic closed form for efficient computation [13]; 3) the Bhattacharyya distance is also analytic based on the Gaussian distribution [1].

By using the Gaussian distribution, $\nabla_{\boldsymbol{\theta}_i} f(\boldsymbol{\theta}_i)$ can be further represented by $\nabla_{\boldsymbol{m}_i} f(\boldsymbol{\theta}_i)$ and $\nabla_{\boldsymbol{\Sigma}_i} f(\boldsymbol{\theta}_i)$, as proposed in NES [13].

$$\nabla_{\boldsymbol{m}_i} f(\boldsymbol{\theta}_i) = \frac{1}{\mu} \sum_{k=1}^{\mu} \boldsymbol{\Sigma}_i^{-1} (\boldsymbol{x}_i^k - \boldsymbol{m}_i) \cdot f(\boldsymbol{x}_i^k)$$

$$\nabla_{\boldsymbol{\Sigma}_i} f(\boldsymbol{\theta}_i) = \frac{1}{\mu} \sum_{k=1}^{\mu} \left( \frac{1}{2} \boldsymbol{\Sigma}_i^{-1} (\boldsymbol{x}_i^k - \boldsymbol{m}_i)(\boldsymbol{x}_i^k - \boldsymbol{m}_i)^{\mathrm{T}} \boldsymbol{\Sigma}_i^{-1} - \frac{1}{2} \boldsymbol{\Sigma}_i^{-1} \right) \cdot f(\boldsymbol{x}_i^k) \quad (13)$$

Similarly, by using the Gaussian distribution, $\nabla_{\boldsymbol{\theta}_i} d(p(\boldsymbol{\theta}_i))$ can be further represented by $\nabla_{\boldsymbol{m}_i} d(p(\boldsymbol{\theta}_i))$ and $\nabla_{\boldsymbol{\Sigma}_i} d(p(\boldsymbol{\theta}_i))$. Given $C(p(\boldsymbol{\theta}_i), p(\boldsymbol{\theta}_j))$ for Gaussian distribution [1], $d(p(\boldsymbol{\theta}_i))$ can be analytically obtained as Eq.(14), $\nabla_{\boldsymbol{m}_i} d(p(\boldsymbol{\theta}_i))$ and $\nabla_{\boldsymbol{\Sigma}_i} d(p(\boldsymbol{\theta}_i))$ can be derived as Eq.(15).

$$d(p(\boldsymbol{\theta}_i)) = \sum_{j=1}^{\lambda} \frac{1}{8} (\boldsymbol{m}_i - \boldsymbol{m}_j)^{\mathrm{T}} \left( \frac{\boldsymbol{\Sigma}_i + \boldsymbol{\Sigma}_j}{2} \right)^{-1} (\boldsymbol{m}_i - \boldsymbol{m}_j) + \frac{1}{2} \log \left( \frac{\left| \frac{\boldsymbol{\Sigma}_i + \boldsymbol{\Sigma}_j}{2} \right|}{\sqrt{|\boldsymbol{\Sigma}_i| \cdot |\boldsymbol{\Sigma}_j|}} \right) \quad (14)$$

$$\nabla_{\boldsymbol{m}_i} d(p(\boldsymbol{\theta}_i)) = \frac{1}{4} \sum_{j=1}^{\lambda} \left( \frac{\boldsymbol{\Sigma}_i + \boldsymbol{\Sigma}_j}{2} \right)^{-1} (\boldsymbol{m}_i - \boldsymbol{m}_j)$$

$$\nabla_{\boldsymbol{\Sigma}_i} d(p(\boldsymbol{\theta}_i)) = \frac{1}{4} \sum_{j=1}^{\lambda} \left( \left( \frac{\boldsymbol{\Sigma}_i + \boldsymbol{\Sigma}_j}{2} \right)^{-1} - \frac{1}{4} \left( \frac{\boldsymbol{\Sigma}_i + \boldsymbol{\Sigma}_j}{2} \right)^{-1} (\boldsymbol{m}_i - \boldsymbol{m}_j)(\boldsymbol{m}_i - \boldsymbol{m}_j)^{\mathrm{T}} \left( \frac{\boldsymbol{\Sigma}_i + \boldsymbol{\Sigma}_j}{2} \right)^{-1} - \boldsymbol{\Sigma}_i^{-1} \right) \quad (15)$$

Thus, $\nabla_{\boldsymbol{m}_i} \mathcal{J}$ and $\nabla_{\boldsymbol{\Sigma}_i} \mathcal{J}$ can be readily obtained by substituting Eqs.(13) and (15) into Eq.(9). Nevertheless, [13] notices that if the above $\nabla_{\boldsymbol{m}_i} \mathcal{J}$ and $\nabla_{\boldsymbol{\Sigma}_i} \mathcal{J}$ are used as the gradients for $\mathcal{J}$, there is a serious issue for directly updating $\boldsymbol{m}_i$ and $\boldsymbol{\Sigma}_i$ with respect to Eq.(10). To be specific, it can be observed that $\nabla_{\boldsymbol{m}_i} \mathcal{J} \propto \frac{1}{\boldsymbol{\Sigma}_i}$ and $\nabla_{\boldsymbol{\Sigma}_i} \mathcal{J} \propto \frac{1}{\boldsymbol{\Sigma}_i^2}$, which means that a large $\boldsymbol{\Sigma}_i$ can make the learning steps of $\boldsymbol{m}_i$ and $\boldsymbol{\Sigma}_i$ insignificant, while a small $\boldsymbol{\Sigma}_i$ can result in a significant update of $\boldsymbol{m}_i$ and $\boldsymbol{\Sigma}_i$. This can lead to an unstable search and thus become impossible to precisely locate the optimum [13]. To address this issue, [13] derives the Fisher information matrix $\mathbf{F}$ from the natural gradient of a population. Here we extend it to the multi-population cases where each pair of $\mathbf{F}_{\boldsymbol{m}_i}$ and $\mathbf{F}_{\boldsymbol{\Sigma}_i}$ is respectively assigned for a sub-population, shown as Eq.(16).

$$\mathbf{F}_{\boldsymbol{m}_i} = \frac{1}{\mu} \sum_{k=1}^{\mu} \boldsymbol{\Sigma}_i^{-1} (\mathbf{x}_i^k - \boldsymbol{m}_i)(\mathbf{x}_i^k - \boldsymbol{m}_i)^{\mathrm{T}} \boldsymbol{\Sigma}_i^{-1}$$

$$\mathbf{F}_{\boldsymbol{\Sigma}_i} = \frac{1}{4\mu} \sum_{k=1}^{\mu} \left( \boldsymbol{\Sigma}_i^{-1} (\mathbf{x}_i^k - \boldsymbol{m}_i)(\mathbf{x}_i^k - \boldsymbol{m}_i)^{\mathrm{T}} \boldsymbol{\Sigma}_i^{-1} - \boldsymbol{\Sigma}_i^{-1} \right) \left( \boldsymbol{\Sigma}_i^{-1} (\mathbf{x}_i^k - \boldsymbol{m}_i)(\mathbf{x}_i^k - \boldsymbol{m}_i)^{\mathrm{T}} \boldsymbol{\Sigma}_i^{-1} - \boldsymbol{\Sigma}_i^{-1} \right)^{\mathrm{T}} \quad (16)$$

With the Fisher information matrix, $\boldsymbol{m}_i$ and $\boldsymbol{\Sigma}_i$ are updated using Eq.(17).

$$\boldsymbol{m}_i = \boldsymbol{m}_i + \eta_m \cdot \mathbf{F}_{\boldsymbol{m}_i}^{-1} \cdot \nabla_{\boldsymbol{m}_i} \mathcal{J}$$

$$\boldsymbol{\Sigma}_i = \boldsymbol{\Sigma}_i + \eta_{\Sigma} \cdot \mathbf{F}_{\boldsymbol{\Sigma}_i}^{-1} \cdot \nabla_{\boldsymbol{\Sigma}_i} \mathcal{J} \quad (17)$$

where $\eta_m$ and $\eta_\Sigma$ are step-size parameters for updating $\mathbf{m}_i$ and $\mathbf{\Sigma}_i$, respectively. Intuitively, since $\mathbf{F}_{\mathbf{m}_i}^{-1} \propto \mathbf{\Sigma}_i^2$ and $\mathbf{F}_{\mathbf{\Sigma}_i}^{-1} \propto \mathbf{\Sigma}_i^4$, it turns out that $\mathbf{F}_{\mathbf{m}_i}^{-1} \cdot \nabla_{\mathbf{m}_i} J \propto \mathbf{\Sigma}_i$ and $\mathbf{F}_{\mathbf{\Sigma}_i}^{-1} \cdot \nabla_{\mathbf{\Sigma}_i} J \propto \mathbf{\Sigma}_i^2$ will no longer oscillate the search.

Notice that, the above equations are computationally intensive. Specially, the inversion of the Fisher matrix subjects to the computational complexity of $O(D^6)$ if the full covariance matrix are considered [13], where $D$ indicates the dimensionality of the search space. To alleviate the computational costs, we simply restrict the covariance matrix and the Fisher matrix for each distribution to be diagonals. This implies that the interdependencies among decision variables are omitted. Although it may make the algorithm less robust to non-separable problems, it suffices to significantly reduce the computational complexity to $O(D)$, as well as to improve the scalability of the algorithm [22].

Another technique adopted from [13] is the normalization of the fitness values. This is motivated by the difficulty of setting a proper trade-off parameter $\varphi$ for aggregating $\nabla_{\boldsymbol{\theta}_i} f(\boldsymbol{\theta}_i)$ and $\nabla_{\boldsymbol{\theta}_i} d(p(\boldsymbol{\theta}_i))$, as different problems may have quite varied scales of fitness values. For that purpose, the utility function in [13] is employed in this paper to reshape the fitness values in each sub-population. Specifically, for each sub-population, all $\mu$ solutions are first ranked based on their fitness values, where $\pi(k)$ indicates the rank of the $k$-th solution. Then the utility function for each $i$-th sub-population, denoted as $U_i$, is carried out to reshape the fitness of each $k$-th solution according to Eq.(18). After that, the utility of each solution is used by replacing the term of $f(\boldsymbol{x}_i^k)$ in Eq.(13).

$$U_i(\pi(k)) = \frac{\max\left(0, \log\left(\frac{\mu}{2}+1\right) - \log(\pi(k))\right)}{\sum_{j=1}^{\mu} \max\left(0, \log\left(\frac{\mu}{2}+1\right) - \log(k)\right)} - \frac{1}{\mu} \qquad (18)$$

The step-size parameters $\eta_m$ and $\eta_\Sigma$ can be either tuned off-line or adjusted during the search. In this paper, the following strategy is used to adjust these two parameters at each iteration.

$$\eta_m \leftarrow \eta_m^{init} \cdot \frac{e - e^{\frac{T_{cur}}{T_{max}}}}{e - 1}$$

$$\eta_\Sigma \leftarrow \eta_\Sigma^{init} \cdot \frac{e - e^{\frac{T_{cur}}{T_{max}}}}{e - 1} \qquad (19)$$

where $T_{max}$ is the total time budget for the whole search and $T_{cur}$ is the consumed budget up to now. $e$ is the natural constant. $\eta_m^{init}$ and $\eta_\Sigma^{init}$ are the initialized values for both step-size parameters, respectively. With Eq.(19), these two step-sizes will decrease over iterations from the initialized values to zero.

So far, all the details have been presented to instantiate an NCS algorithm. To summarize, the proposed algorithm is a multi-Gaussian distribution based EA; Each distribution drives the evolution of one sub-population with the well-established NES; Multiple Gaussian distributions are driven to be negatively correlated by the proposed diversity model. In this regard, the proposed algorithm can also be regarded as a new variant of NES that has the ability of parallel exploration. Thus, it is named Negatively Correlated Natural Evolution Strategies (NCNES) for intuition. The detailed steps of NCNES is listed in Algorithm II for reference.

---

**Algorithm II: The proposed NCNES**

1  **Input:** $f$, $\lambda$, $\mu$, $\eta_m^{init}$, $\eta_\Sigma^{init}$, $\varphi$, $T_{max}$
2  **Begin:**
3      For $i = 1$ to $\lambda$:
4          Initialize a Gaussian distribution for the $i$-th Search Process as $\mathcal{N}(\boldsymbol{m}_i, \boldsymbol{\Sigma}_i)$;
5      $T_{cur} = 0$;
6      **While** $T_{cur} < T_{max}$ **do:**
7          $\eta_m \leftarrow \eta_m^{init} \cdot \frac{e - e^{\frac{T_{cur}}{T_{max}}}}{e - 1}$;
8          $\eta_\Sigma \leftarrow \eta_\Sigma^{init} \cdot \frac{e - e^{\frac{T_{cur}}{T_{max}}}}{e - 1}$;
9          For $i = 1$ to $\lambda$:
10             Generate $\mu$ solutions $\boldsymbol{x}_i^k \leftarrow \mathcal{N}(\boldsymbol{m}_i, \boldsymbol{\Sigma}_i)$, $\forall k = 1, \dots, \mu$;
11             Evaluate the fitness $f(\boldsymbol{x}_i^k)$, $\forall k = 1, \dots, \mu$;
12             $T_{cur} \leftarrow T_{cur} + \mu$;
13             Update $\boldsymbol{x}^*$ as the best solution ever found;
14             Rank the $k$-th solution in terms of its fitness $f(\boldsymbol{x}_i^k)$ as $\pi(k)$, $\forall k = 1, \dots, \mu$;
15             Set $U_i(\pi(k)) = \frac{\max(0, \log(\frac{\mu}{2}+1) - \log(\pi(k)))}{\sum_{j=1}^{\mu} \max(0, \log(\frac{\mu}{2}+1) - \log(k))} - \frac{1}{\mu}$, $\forall k = 1, \dots, \mu$;
16             $\nabla_{\boldsymbol{m}_i} f \leftarrow \frac{1}{\mu} \sum_{k=1}^{\mu} \boldsymbol{\Sigma}_i^{-1} (\boldsymbol{x}_i^k - \boldsymbol{m}_i) \cdot U_i(\pi(k))$;
17             $\nabla_{\boldsymbol{\Sigma}_i} f \leftarrow \frac{1}{2\mu} \sum_{k=1}^{\mu} (\boldsymbol{\Sigma}_i^{-1}(\boldsymbol{x}_i^k - \boldsymbol{m}_i)(\boldsymbol{x}_i^k - \boldsymbol{m}_i)^{\mathrm{T}} \boldsymbol{\Sigma}_i^{-1} - \boldsymbol{\Sigma}_i^{-1}) \cdot U_i(\pi(k))$;
18             $\nabla_{\boldsymbol{m}_i} d \leftarrow \frac{1}{4} \sum_{j=1}^{\lambda} \left(\frac{\boldsymbol{\Sigma}_i + \boldsymbol{\Sigma}_j}{2}\right)^{-1} (\boldsymbol{m}_i - \boldsymbol{m}_j)$;
19             $\nabla_{\boldsymbol{\Sigma}_i} d \leftarrow \frac{1}{4} \sum_{j=1}^{\lambda} \left(\left(\frac{\boldsymbol{\Sigma}_i + \boldsymbol{\Sigma}_j}{2}\right)^{-1} - \frac{1}{4}\left(\frac{\boldsymbol{\Sigma}_i + \boldsymbol{\Sigma}_j}{2}\right)^{-1} (\boldsymbol{m}_i - \boldsymbol{m}_j)(\boldsymbol{m}_i - \boldsymbol{m}_j)^{\mathrm{T}} \left(\frac{\boldsymbol{\Sigma}_i + \boldsymbol{\Sigma}_j}{2}\right)^{-1} - \boldsymbol{\Sigma}_i^{-1}\right)$;
20             $\mathbf{F}_{\boldsymbol{m}_i} \leftarrow \frac{1}{\mu} \sum_{k=1}^{\mu} \boldsymbol{\Sigma}_i^{-1} (\boldsymbol{x}_i^k - \boldsymbol{m}_i)(\boldsymbol{x}_i^k - \boldsymbol{m}_i)^{\mathrm{T}} \boldsymbol{\Sigma}_i^{-1}$;
21             $\mathbf{F}_{\boldsymbol{\Sigma}_i} \leftarrow \frac{1}{4\mu} \sum_{k=1}^{\mu} \left(\boldsymbol{\Sigma}_i^{-1}(\boldsymbol{x}_i^k - \boldsymbol{m}_i)(\boldsymbol{x}_i^k - \boldsymbol{m}_i)^{\mathrm{T}} \boldsymbol{\Sigma}_i^{-1} - \boldsymbol{\Sigma}_i^{-1}\right) \left(\boldsymbol{\Sigma}_i^{-1}(\boldsymbol{x}_i^k - \boldsymbol{m}_i)(\boldsymbol{x}_i^k - \boldsymbol{m}_i)^{\mathrm{T}} \boldsymbol{\Sigma}_i^{-1} - \boldsymbol{\Sigma}_i^{-1}\right)^{\mathrm{T}}$;
22             $\boldsymbol{m}_i \leftarrow \boldsymbol{m}_i + \eta_m \cdot \mathbf{F}_{\boldsymbol{m}_i}^{-1} (\nabla_{\boldsymbol{m}_i} f + \varphi \cdot \nabla_{\boldsymbol{m}_i} d)$;
23             $\boldsymbol{\Sigma}_i \leftarrow \boldsymbol{\Sigma}_i + \eta_\Sigma \cdot \mathbf{F}_{\boldsymbol{\Sigma}_i}^{-1} (\nabla_{\boldsymbol{\Sigma}_i} f + \varphi \cdot \nabla_{\boldsymbol{\Sigma}_i} d)$;
24 **Output** $\boldsymbol{x}^*$, $f(\boldsymbol{x}^*)$.

# Section IV NCNES for Reinforcement Learning

EAs are intuitively promising solutions to Reinforcement Learning (RL) problems as the population-based nature of EAs not only provides the urgent exploration ability to RL [25], but also provides other merits such as parallel acceleration [33], noisy-resistance [34], and compatibility of training non-differentiable policies (e.g., trees [35]). For example, the canonical NES has been successfully shown to be a promising reinforcement learning method by playing Atari games [25]. Furthermore, RL problems are naturally good testbeds for NCNES as the performance of solving RL problems is highly dependent on the exploration ability when optimizing the policy [32]. On this basis, this paper empirically studies the NCNES-based solution for Reinforcement Learning (RL) problems by playing Atari games.

For the purpose of performance assessment, the empirical studies will uncover three-fold advantages about how effectively the new NCS framework facilitates the search, how well the proposed new diversity model contributes to NCNES, and how well NCNES behaves on reinforcement learning problems.

**Section IV.A Reinforcement Learning**

RL is a class of problems that learns to make Markov decisions so that the long-term rewards can be maximized. In RL, the policy can be iteratively learnt only by interacting with the environment. At each time step, the agent picks an action according to the policy and the observed state of environment, leading to a transition from origin state to the next state, then receives a reward as the feedback to update the policy. The above steps keep going until terminated. To maximize the expected cumulative discounted reward in long term, numerous RL methods have been developed in the last decades, e.g., the model-based methods [26][27], the value function based methods [28][29], and the policy search based methods [25][30]. For more details of RL methods, please refer to [31].

Among the existing works, the policy search based methods that adopt the deep neural networks as the policy model have drawn most research attentions due to their powerful performance [25][30]. The key problem for this type of methods turns into how to train the deep network in the RL settings, which faces three major difficulties. First, the search space of training the deep neural networks is highly large-scale and multi-modal; second, due to the Markov decision process nature of RL, the policy learning process is non-differentiable unless some derivable functions are specially designed (e.g., the critic function in A3C [30]); last, the delayed rewards may involve considerable noise. NES is a suitable method for learning the policy due to its derivate-free, robust and parallel features. Empirical studies on a set of Atari games have verified the advantages of NES over several state-of-the-art methods [25].

**Section IV.B NCNES for Playing Atari**

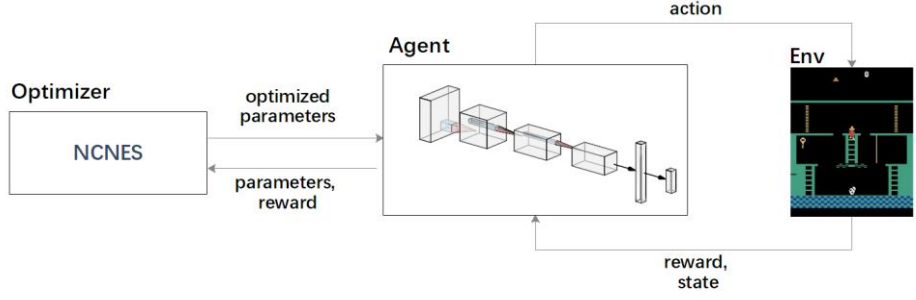

Fig. 1 The flowchart of NCNES based solution for playing Atari

Atari 2600 is a set of video games that have been popular for over 40 years. Atari games successfully cover different types of difficult tasks, such as obstacle avoidance (e.g., Freeway and Enduro), shooting (Beamrider) and other types. The player can do various actions in each game so as to maximize the cumulative reward solely by interacting with the game environment. Due to these features, Atari games have been widely used as the RL simulation platform for empirical studies.

The flowchart of applying NCNES to play Atari games can be seen in Fig.1 for illustration. Basically, the agent aims to learn the policy by iteratively imposing actions to the Atari environment and getting states and rewards therefrom. The policy is modeled as a deep convolutional network for the purpose of conveniently and effectively processing the high-dimensional raw pixel data that is directly received from the video games. NCNES is applied to optimize the connection weights of the policy network without back-propagation. The network architecture of the agent consists of three convolution layers and two full connection layers (see Table I), as suggested by [28].

Table I : The Network Architecture of the Agent

|        | Input     | Output     | Kernel Size | Stride | #filters | activation |
|--------|-----------|------------|-------------|--------|----------|------------|
| **Conv1** | 4x84x84   | 32x20x20   | 8x8         | 4      | 32       | ReLU       |
| **Conv2** | 32x20x20  | 64x9x9     | 4x4         | 2      | 64       | ReLU       |
| **Conv3** | 64x9x9    | 64x7x7     | 3x3         | 1      | 64       | ReLU       |
| **Fc1**   | 64x7x7    | 512        | -           | -      | -        | ReLU       |
| **Fc2**   | 512       | Action num | -           | -      | -        | -          |

More specifically, each individual solution is represented as a vector of all the connection weights of the policy model. Accordingly, the distributions of NCNES search processes are estimated based on those high-dimensional solutions. The training phase is divided into multiple epochs. At each epoch, the agent starts from the beginning of the game and takes a sequence of actions from the policy model to react to the environment, so as to gain as many as possible scores until game overs. After a game (i.e., an epoch) has been finished, the reward will be returned back to the agent as well as NCNES. Then NCNES takes the reward of each epoch as the fitness value of each iteration to optimize the connection weights (generating a population of new policy models for the next epoch) in a parallel exploration way, i.e., together with diversity among different search processes. When the training budget runs out, the final policy model will be output for further usages.

From the perspective of optimization, the above problem-solving procedure suffers from three kinds of difficulties.
- First, the search space is extremely large-scale. The deep architecture of the policy results in huge numbers of connection weights to be optimized, where NCNES needs to solve 1.7 million dimensional real-valued optimization problems.
- Second, the search space is highly multi-modal due to the complex architecture of the deep neural networks and the non-uniform distribution of the rewards.
- Third, the feedback is quite uncertain. On one hand, the reward is heavily delayed as the agent can only get the total reward from the environment after the game playing is ended, which makes it very difficult to evaluate the subtle action at each timestep of an epoch. On the other, the total reward involves considerable noise introduced by the randomized Atari games settings, which makes it even harder to evaluate the policy.

Due to the large-scale, uncertain, and multi-modal nature, the optimization problem is non-trivial at all.

**Section IV.C Experimental Protocol**

Three Atari games are selected for the empirical studies, i.e., Freeway, Enduro, and Beamrider. The screenshots of these three games are shown in Fig.2. In freeway, the pedestrian is controlled by three actions (up, down and wait), aiming at avoiding dangerous collisions when goes across a ten-lane highway with large traffic volume, and scores every time it succeeds to reach the other side. The player in Enduro maneuvers a race car to avoid other racers and achieves higher mileage in an endurance race last for several "days" (counted in the game). The decreased visibility in night or severe weather, and the increased car speed as well as the frequency have posted great challenges. Beamrider is a horizontal scrolling short-range shooter targeted at shooting off destroyable coming enemies with a limited supply of torpedoes and escaping from other undefeatable enemies.

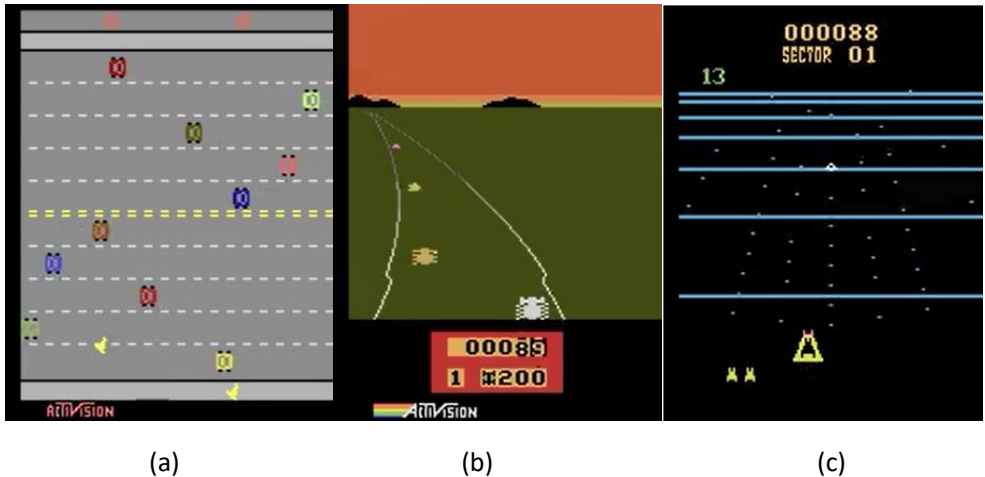

(a)　　　　　　　　　　(b)　　　　　　　　　　(c)

Fig. 2 The screenshots of these three games, where (a) is for the Freeway, (b) is for the Enduro and (c) is for the Beamrider.

Three RL methods are selected for comparisons, denoted as A3C [30], CES [25] and NCS-C [1], respectively. All those methods are incorporated into the policy search based RL framework for training the same deep neural network as NCNES does, i.e., optimizing the connection weights. Among them, A3C is a state-of-the-art gradient-based method that trains the network with the traditional back-propagation. The other two algorithms are EA-based optimization method. CES is the canonical NES that has been successfully applied to play the Atari games [25]. NCS-C is the instantiation of the original NCS framework. Both the well-established A3C and CES can be used to demonstrate the effectiveness of NCNES on playing Atari games. CES can also be used to assess how parallel exploration can facilitate the search, as NCNES can be simply viewed as a new variant NES with parallel exploration ability. NCS-C is used to show the advantages of the proposed new NCS framework over the original NCS.

For all the comparisons, each algorithm terminates the training phase in a game when the total time budget runs out, and the final solution (policy network) will be returned for testing. The quality of the final solution is measured with the testing score, i.e., averaged score of 30 repeated run in one game-playing without the time limitations. Considering that the environment of a game-playing is randomly initialized, each game-playing will be repeated for three times, i.e., there will be three testing scores for each algorithm on each game. The total time budget is set as the total game frames that each algorithm is allowed to consume for training. For three EA-based methods, the total game frames are set to 100 million. For A3C, as it works quite differently with back-propagation, it is unfair to set the same total game frames with the EA-based ones. In this regard, we counted the game frames consumed by both well-established CES and A3C on the same hardware conditions and in the same game with the same given computational run time. It has been found that the ratio of the consumed game frames between them is about 2.5. As a result, the total game frames are set to 40 million for A3C for fairness. To discretize the games for agent's actions execution and states acquiring, the skipping frame is set to 4. That is, for each training phase, the agent is allowed to take 25 million actions for EA-based methods and 10 million actions for gradient-based method.

As both CES and A3C have been successfully applied to play Atari games, we directly borrow the hyperparameters settings from the corresponding papers [25][30]. The hyperparameters of both NCS-C and NCNES are given as follows. For NCS-C, the number of search processes is set to 8, the sigma is initialized to 0.01, the learning rate of the sigma and the learning epoch are set the same with its original paper [1]. To reduce the noise of the environment, each solution will be re-evaluated for 10 times at each epoch of the training phase, and the averaged score will be returned to NCS-C as the fitness for the solution. For NCNES, the hyperparameters are listed in Table II for brevity.

**Table II The Hyperparameter Settings of NCNES**

| Parameter | Value | Remark |
|---|---|---|
| $\lambda$ | 5 | The number of sub-populations |
| $\mu$ | 15 | The individuals in each sub-population |
| $\varphi$ | 0.0001 | The trade-off parameter for balancing the exploration and exploitation, set based on the statistically approximated ratio between the scales of fitness gradients and the diversity gradients. |
| $\eta_m^{init}$ | 0.5 | The initial learning rate of mean vectors |
| $\eta_n^{init}$ | 0.1 | The initial learning rate of covariance matrix |

| | | | |
|---|---|---|---|
| $t$ | Randomly pick from [1,2,3,4,5] | The re-evaluation times of each solution to reduce the environmental uncertainty | |

### Section IV.D Results and Analysis

***Performance Analysis on Game Scoring.*** Three repeated testing scores of each algorithm on three games are shown in Table III. It can be clearly seen that, NCNES can outperform all the compared algorithms on the tested three games, which successfully verifies the effectiveness of NCNES on reinforcement learning problems. By comparing NCNES with CES, it suffices to show that the parallel exploration can facilitate the search much better as NCNES gains averagely twice scores over CES. By comparing NCNES with NCS-C, it can be seen that NCNES gains around three times scores over NCS-C on Freeway, and NCNES also shows significant advantages on other two games. This verifies the effectiveness of the mathematical NCS model. A3C performs less robust than the other three algorithms as its final policy model fails to gain any scores in two games. This maybe because the population-based search can reduce the uncertainty of the algorithms themselves, by 1) frequently sampling from a small region of the search space, which plays the role of re-evaluations to some extent; 2) only requiring the relative order of solutions to determine the search direction, which is less sensitive to the evaluation noise.

Table III The averaged testing scores of four algorithms on three Atari games

| Game | | CES | A3C | NCS-C | NCNES |
|---|---|---|---|---|---|
| Game Frames | | 100M | 40M | 100M | 100M |
| Freeway | Run 1 | 15.9 | 0.0 | 7.0 | **22.7** |
| | Run 2 | 12.7 | 0.0 | 9.4 | **21.1** |
| | Run 3 | 14.1 | 0.0 | 3.7 | **22.1** |
| | Average | 14.23 | 0.0 | 6.7 | **22.0** |
| Beamrider | Run 1 | 401.0 | 908.0 | 602.0 | **856.8** |
| | Run 2 | 508.2 | 490.2 | 686.0 | **620.4** |
| | Run 3 | 414.1 | 336.0 | 482.0 | **719.3** |
| | Average | 441.1 | 646.7 | 590.0 | **732.2** |
| Enduro | Run 1 | 6.2 | 0.0 | 6.0 | **29.8** |
| | Run 2 | 7.0 | 0.0 | 12.8 | **8.7** |
| | Run 3 | 8.1 | 0.0 | 6.4 | **11.5** |
| | Average | 7.1 | 0.0 | 8.4 | **17.9** |

***Performance Analysis on Convergence Speed.*** To study from the optimization perspective, the score curves of four algorithms on three games are depicted in Fig.3. To depict the curves, at the end of each epoch of the training phase, the current best policy model in terms of the training scores, will be additionally tested for 30 times. And the averaged testing score will be recorded for the purpose of depicting the score curve. Note that, this testing time will not be counted into the total game frames budget, as this score will not be used for helping training. Then the testing score is depicted epoch-by-epoch to form the score curve. Generally, the score curve of an algorithm can express the convergence speed of the optimization algorithm. It can be seen that, NCNES (the red curve) can usually search a very good policy model in very short timesteps. This means that even with a much smaller time budget, NCNES can still outperform the others. For NES and NCS-C, the score curves increase much slower along with the timesteps. This verifies that the new NCS model can facilitate the search more effectively. Although A3C can occasionally gain high scores, it is very unstable as the score curves oscillate heavily, which even returns very bad policy models (i.e., the averaged score is 0.0 for two games) as the final output. This might be that A3C is less resistant to the environmental noise.

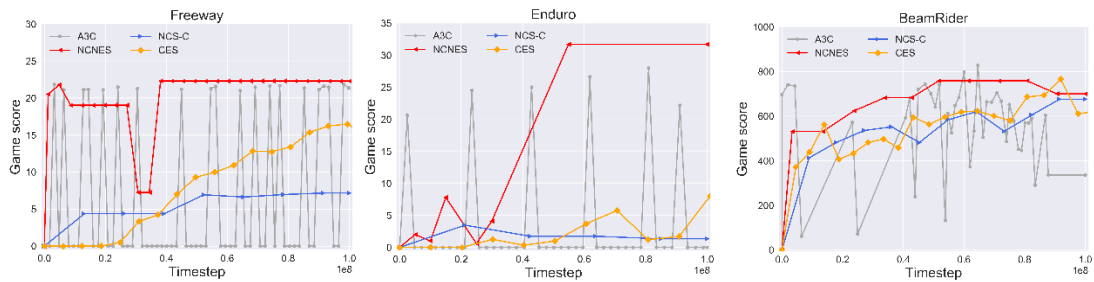

Fig.3 The score curves of four algorithms on three games, respectively.

*Performance Analysis on Policy Behaviors*. It is expected that the parallel exploration search behavior of NCNES can help emerge some novel yet useful behaviors that traditional policies are less likely to express. For BeamRider, the agent trained by NCNES prefers staying in the left side of the available area and gains as many as 996 scores in a single testing play (see Fig.4). The motivation behind this trick can be explained as that staying in the left side can prevent at most 50% enemy attacks, and thus is beneficial to longer survival. For Enduro, the agent prefers driving in the middle of the racing track when the weather is good so as to preserve the maximal freedom to move to both sides (see Fig.5(a)). When the visibility decreases as it is snowy, foggy, and night, the agent prefers driving at one side of the racing track for safety, similar to human behaviors (see Figs.5 (b)-(d)).

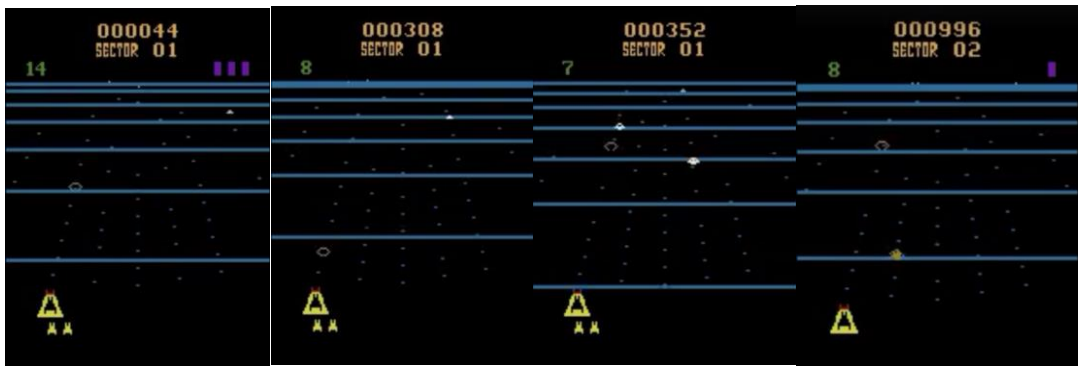

Fig.4 Tricks learned in BeamRider: the agent prefers staying in the left side of the available area.

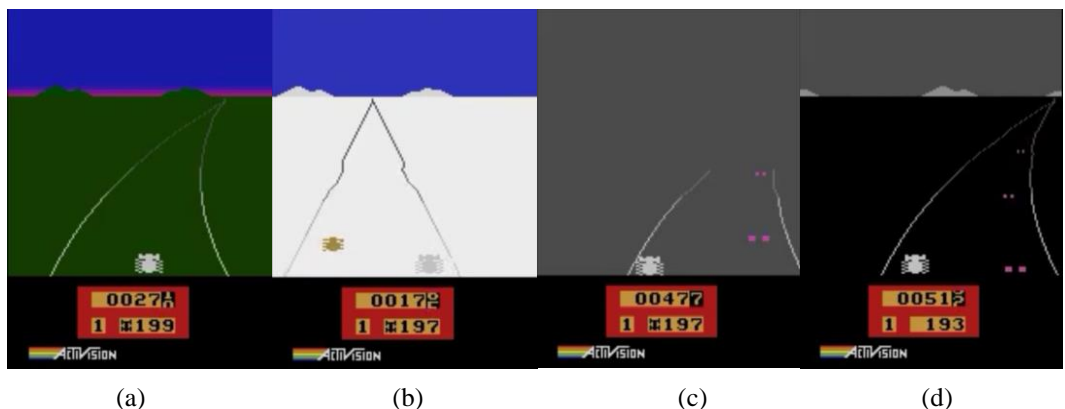

Fig.5 Tricks learned in Enduro: the agent prefers (a) driving in the middle of the racing track when the weather is good so as to preserve the maximal freedom to move to both sides. When the visibility decreases as it is (b) snowy, (c) foggy, and (d) night, the agent prefers driving at one side of the racing track for safety.

***Performance Analysis Parallel Acceleration.*** Lastly, we show how NCNES can utilize the parallel computing resources. To be specific, three kinds of NCNES are implemented. For the first kind (see Fig.6(a)), NCNES is run on one computing unit in a serial manner. For the second kind (see Fig.6(b)), NCNES is implemented in an island-model based architecture, i.e., 5 search processes are run on 5 fixed computing cores respectively during the whole optimization process; At each iteration, information transferring among computing units only happens when the diversity gradients are calculated (also see Algorithm 1, step 10). For the third kind (see Fig.6(c)), NCNES is implemented in a hybrid architecture; Specifically, 5 search processes are run in an island-model manner with 5 groups of computing units, each group is organized in a master-slave model that consists of 15 computing cores for the fitness evaluations of 15 individuals of a search process, respectively. All three implementations of NCNES are independently run on the same workstation (Intel(R) Xeon(R) CPU E5-2699A v4 @ 2.40GHz) with 44 cores (88 threads)[3].

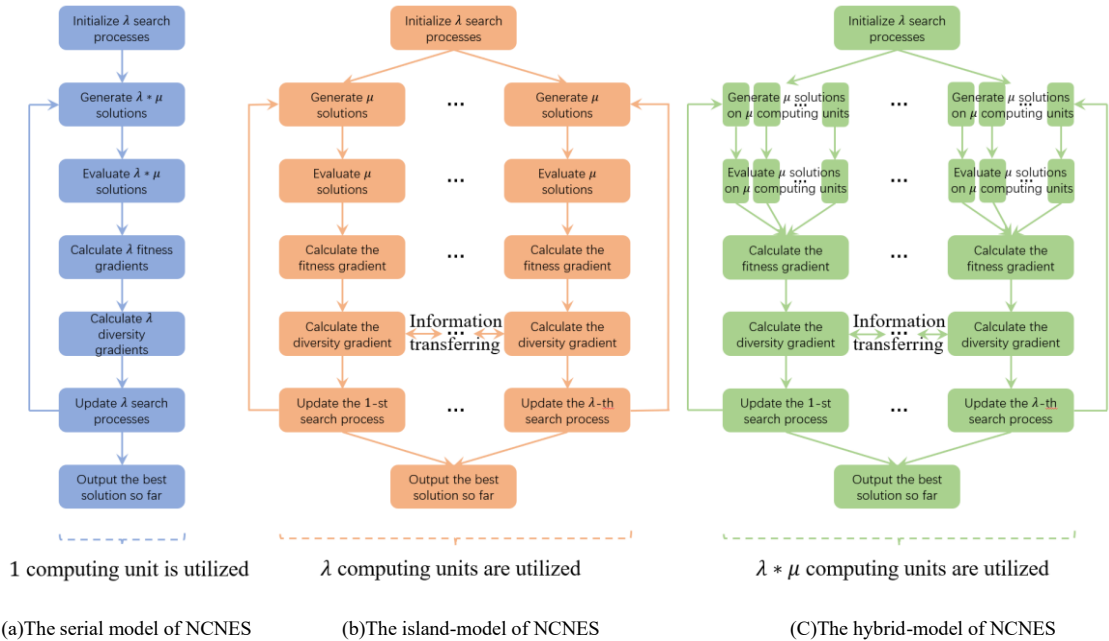

Fig. 6 The flowcharts of three kinds of implementations for NCNES.

The above three implementations of NCNES are simulated on three games with 100M training frames., where three independent runs are conducted for each game. The runtime results are listed in Table IV. It can be seen that, both island-model and hybrid-model can effectively utilize the parallel computing resources for acceleration. That is, by running on a common lab-level workstation, the computing runtime can be largely reduced from around 120 hours (by the serial model) to as short as 2 hours (by the hybrid model). Furthermore, given that the population size of NCNES (including both $\lambda$ and $\mu$) can be easily enlarged to enhance the parallel exploration ability, it would be interesting to assess how NCNES can be speedup with large-scale computing resources. This can be measured with the speedup ratio. Theoretically, the speedup ratio [4] $r \in [0,1]$ says that given $m$ computing units, the parallel implementation can reduce the runtime for $m * r$ times. The speedup ratios of both island-model and hybrid-model on Freeway and Enduro are very promising, i.e., stably above 0.72. On the other hand, the speedup ratio on Beamrider is unsatisfactory, i.e., 0.43 for island-model and 0.09 for hybrid model. Actually, this is mostly caused by the blocking synchronization used for transferring the distribution parameters while calculating the diversity gradients. Specifically, at each iteration of NCNES, the solutions are re-evaluated by playing multiple times of the game. As the durations of each game playing can be various for different solutions (e.g., from minutes to hours for Beamrider), each search process may reach the information transferring step at quite different timesteps. Unfortunately, the blocking synchronization would calculate the diversity gradients unless all the distribution parameters are received by each search process. Fortunately, this waiting time can be greatly eliminated by employing the non-blocking asynchronization for approximate information transferring, since the distribution parameters to be transferred has already been obtained at the previous iteration (see Algorithm 1, step 12) and can be transferred at any time afterwards. The price to pay would be the accuracy of the information transferring. To summarize, due to the parallel exploration feature, NCNES is able to be effectively accelerated by parallel computing resources if the computational loads can be well balanced.

---

[3] The tasks of calculating the diversity gradients (as well as fitness gradients calculation and search processes updating) are always fixed to different physical cores, otherwise the memory sharing mechanism may influence the information transferring efficiency.

**Table IV The simulated computing runtime of three implementations of NCNES**

| Computing model | | Serial model | Island-model | Hybrid-model |
|---|---|---|---|---|
| Computing Units (i.e., $m$) | | 1 | 5 | 75 |
| Computing Runtime | Freeway | 122.6±0.5 hours | 31.2±0.2 hours | 2.28±0.0 hours |
| | BeamRider | 116.0±18.8 hours | 58.8±22.2 hours | 19.48±4.3 hours |
| | Enduro | 119.6±0.7 hours | 30.4±0.1 hours | 2.16±0.0 hours |
| Speedup Ratio ($\frac{\text{runtime}_{serial}}{\text{runtime}_{parallel} \times m}$) | Freeway | - | 0.78±0.01 | 0.72±0.00 |
| | BeamRider | - | 0.43±0.20 | 0.09±0.03 |
| | Enduro | - | 0.79±0.01 | 0.74±0.02 |

## Section V Conclusions

In this paper, we propose a new mathematically principled NCS framework. The new NCS works by explicitly modeling and maximizing the diversity model (for exploration) and the fitness model (for exploitation) of the next population. Both models can be maximized through gradient descending with respect to each search process. Comparing to the original NCS, the new NCS has clearer mathematical explanations of why the negatively correlated search processes can lead to a parallel exploration search behavior and how to optimally realize it. Besides, the new NCS has also successfully addressed two technical issues of the original NCS. To assess the performance of the new NCS, an instantiation called NCNES is presented. NCNES adopts the well-established NES as the search strategy of each sub-population. NCNES is then applied to solve RL problems for playing Atari games. Specially, NCNES is used to directly train a set of 1.7 million connection weights of the deep policy model under various uncertainties. Empirical studies have shown that, on three typical Atari games, NCNES is able to significantly outperform the state-of-the-arts methods (including both EA-based solution and gradient-based solution). By pairwise comparisons, it also verifies that the proposed new NCS model is better than the original NCS for the purpose of parallel exploration, and the parallel exploration ability can facilitate the search performance as well as the computational efficiency of the new NCS.

---

[4] The speedup ratio is measured as the ratio of the accumulated runtime on all computing units between the serial implementation and the parallel implementation. Suppose the serial model uses 1 computing unit and its runtime is denoted as $runtime_{serial}$, and the parallel model uses $m$ computing units and the runtime is denoted as $runtime_{parallel}$, then the speedup ratio is calculated as $\frac{runtime_{serial}}{runtime_{parallel} \times m}$. The theoretical speedup ratio $r$ varies from 0.0 to 1.0, where $r = 1.0$ indicates the optimal linear speedup. Though some techniques like memory sharing can practically boost $r$ over 1.0, they are avoided in this work as the footnote 3 mentioned.